# Engineering FAIR Privacy-preserving Applications that Learn Histories of Disease

**Inês N. Duarte[1,2,3], Praphulla M. S. Bhawsar[1], Lee K. Mason[1], Jeya Balaji Balasubramanian[1], Daniel E. Russ[1], Arlindo L. Oliveira[2,3], Jonas S. Almeida[1]**

[1]Division of Cancer Epidemiology and Genetics, National Cancer Institute, National Institutes of Health, Rockville, MD, United States; [2]INESC-ID, R. Alves Redol 9, Lisbon 1000-029, Portugal; [3]Instituto Superior Técnico, Universidade de Lisboa, Av. Rovisco Pais, Lisbon 1049-001, Portugal

**Abstract**

*A recent report on "Learning the natural history of human disease with generative transformers" created an opportunity to assess the engineering challenge of delivering user-facing Generative AI applications in privacy-sensitive domains. The application of these models, particularly for personalized healthcare tasks like predicting individual morbidity risk, is typically constrained by data privacy concerns. This project was accordingly designed as an in-browser model deployment exercise (an "App") testing the architectural boundaries of client-side inference generation (no downloads or installations). We relied exclusively on the documentation provided in the reference report to develop the model, specifically testing the "R" component of the FAIR data principles: Findability, Accessibility, Interoperability, and Reusability. The successful model deployment, leveraging ONNX and a custom JavaScript SDK, establishes a secure, high-performance architectural blueprint for the future of private generative AI in medicine.*

1. **Introduction**

Predictive modeling of health trajectories[1,2] holds immense potential for personalized medicine[3–6]. The ability of generative models, such as Delphi 2M[7], to learn the natural history of disease[8,9] from electronic health records[10] and predict future morbidity or mortality events[3] is a powerful tool for early intervention and individualized care. However, the application of these models introduces critical challenges related to privacy, governance, and technological accessibility. Using generative transformers, particularly for sensitive personalized healthcare tasks, typically requires sending patient health records to powerful, centralized cloud servers for processing. This creates significant regulatory and ethical challenges. Our work addresses this critical gap by transforming the Delphi 2M model into an artifact that can be executed entirely within a standard web browser. The original project[7] provides the code and a synthetic dataset to train a version of the model. In addition, we use web technologies to specifically improve the Reusability of the inference pipeline (the R component of the FAIR data principles)[11–13].

The FAIR principles[14] guide the management of scientific data and resources[15] to maximize their value. In this context, the original Delphi 2M model, while well-documented, is primarily designed for conventional cloud or institutional deployment, making the inference utility difficult for a user to access without submitting sensitive data. Our work shortcuts this challenge by providing a completely client-side application, eliminating the need for data upload or application installation. Specifically, we address Interoperability (I) by making use of the Open Neural Network Exchange (ONNX) format[16]. This open standard acts as a universal interchange format, enabling the model to be understood and executed across different runtimes (in our case, JavaScript/WebAssembly)[16]. Our primary engineering objective was to improve the Interoperability and Reusability of the model logic and parameters from the original research. This was achieved by successfully migrating the entire inference pipeline from the original Python to a client-side JavaScript environment[11].

Ultimately, the feasibility of this engineering exercise is a testament of the importance of ONNX as a shared runtime (Figure 1). This architectural choice aligns the technological solution with strict data governance requirements by placing the computational boundary and control entirely on the user's local device.

## 2. Methods

The starting point for this engineering project was the publicly available source code[17] from the Delphi 2M authors, specifically the training and modeling scripts that define the transformer's architecture and logic. The core implementation, provided as a PyTorch script based on the nanoGPT framework[7], was utilized for two primary reasons: to ensure a high-fidelity capture of the model's learned parameters and, to fully reverse-engineer the required data preprocessing and postprocessing steps necessary for client-side inferencing.

This initial source model, trained in 7144 patients using the provided train.py script[17], and validated in a dataset with 7144 patients, implements a customized Generative Pre-trained Transformer (GPT) architecture. The training process uses a dual loss function to learn both the next medical event and the time until that event occurs. The inference pipeline relies on the management of patient trajectories, tokenization, age encoding, and a custom time-to-event mechanism. We migrated this system from a server-side environment (PyTorch/GPU) into a client-side environment (JavaScript/ONNX).

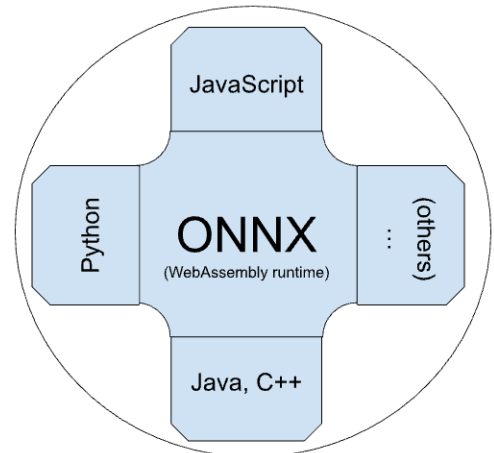

**Figure 1.** ONNX Runtime with interchanging SDKs in multiple languages, see https://onnxruntime.ai.

**Open Neural Network Exchange (ONNX)**

The original Delphi 2M model was developed and trained using the PyTorch framework[18]. To enable execution in resource-constrained and diverse environments such as the browser, we first converted the model to the ONNX format [16].

ONNX[19] serves as an open standard that defines a common set of operators and file formats for representing deep learning models (Figure 1) effectively decoupling the trained model from its original training framework (like PyTorch or TensorFlow) to produce artifacts that can be used in Python, Java, C++, Julia, and crucially on the web.

The modern Web stack supports running code in multiple languages by compiling to an efficient intermediate representation called WebAssembly (Wasm)[20,21]. Wasm is a low-level, binary instruction format that is designed to be a portable compilation target for high-level programming languages like C/C++, enabling deployment on the web. By using the C++ core of the ONNX Runtime compiled into Wasm, we can run complex model inference pipelines in the browser at near-native speeds. The ONNX Runtime Web uses Wasm for CPU-bound computation and WebGPU backends for GPU acceleration when available, providing a viable, high-performance path for client-side deep learning.

In the context of this project, ONNX and its runtime were essential for two primary reasons: first and foremost, portability, as it decouples the model from its original training framework; and secondly efficiency, as the ONNX runtime can be configured to use highly optimized backends like WebGPU, WebGL or Wasm.

For runtime interaction, the SDK manages the low-level communication with the ONNX Runtime Web. Its responsibilities include *loading* the model by calling the runtime to fetch the ONNX file and establish the inference session. The crucial step of tensor creation then takes place, where the SDK converts the user's raw, human-readable inputs into the structured, numeric tensors that precisely match the data types and shapes required by the ONNX model's input signature. Once the inputs are ready, the SDK initiates the execution by passing these tensors to the Runtime's Wasm engine for computation. Finally, after the Runtime executes the inference graph and returns the raw output tensors (logits), the SDK performs Postprocessing, converting these results back into human-readable morbidity risk estimates (events and ages in years) for display to the user.

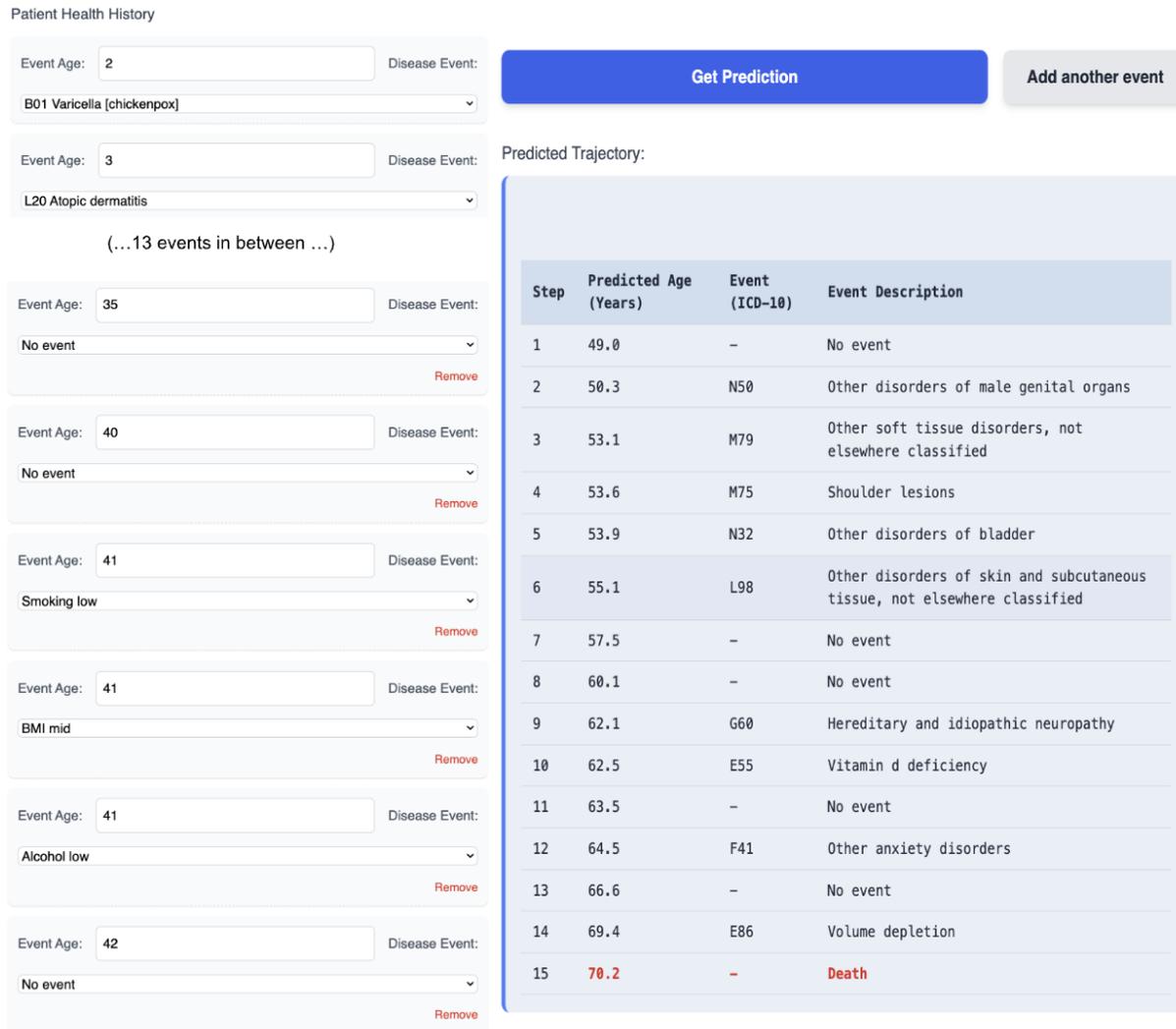

**Figure 2.** Snapshot of the Delphi App in the web browser. The execution of the Delphi 2M model entirely client-side is driven by three critical components: the model format (ONNX), the interface layer (SDK), and a web development environment (ObservableHQ). In this example, the ICD-10 health trajectory of a 42 year old individual (left panel) is followed by a prediction of future trajectories (right panel).

**The Development Pipeline**

The engineering process followed a defined sequence of steps towards a final in-browser functionality. The process begins with Model Conversion where the pre-trained Delphi 2M model is exported from its native framework into the ONNX format, ensuring compatibility with the in-browser ONNX runtime. The next step, SDK Implementation, involves writing a dedicated JavaScript library (the SDK, Figure 1, 3) to wrap the ONNX Runtime Web API, including the essential data preprocessing and postprocessing logic.
The core of the SDK's functionality, the *generateTrajectory* function, performs iterative inference. First, it sends the current event sequence to the ONNX model via *getLogits*, which returns logits representing the probability for each possible subsequent event. This SDK employs a time-to-event sampling method, transforming the *logits* into random time points $t_{sample}$ until the next event, for each ICD-10 medical code, using the formula:

$$t_{sample} = -e^{-logit}.ln(u)$$

where $u$ is a random number generated by a pseudo-random number generator. The event with the minimum predicted time $t_{min}$ is then selected as the next predicted event, and the patient's age is updated by adding $t_{min}$. This iterative loop continues until a termination token is encountered or the generated trajectory exceeds the maximum age. The termination token is set to "Death" and the maximum age to 85 years by default, corresponding to the original Delphi code, but both are parameters that can be set by the user of the SDK.

The final step involved bringing the finalized SDK into the main exploratory web application (Figure 3), forming the backbone of the client-side risk estimation feature, which manages user input and displays real-time predictions without server communication for inference.

To conclude the study, we developed a web application (Figure 2) that allows non-technical users to interact with the system without software installation. The interface is designed to reflect a clinical timeline. The upper section accepts the input event sequence, while the lower section visualizes the calculated trajectory as it is generated. This validates the ability to operate these workflows directly in the user's web browser, and can be put to the test at https://epiverse.github.io/delphiTrajectories.

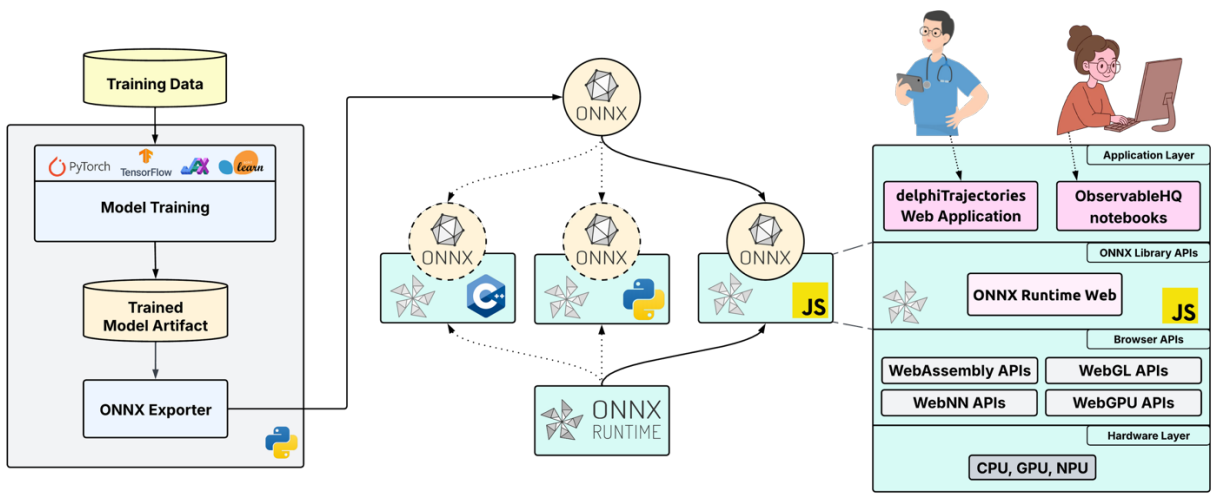

Figure 3. End-to-end engineering pipeline from data to browser. The diagram shows how the original PyTorch model[17] is converted into an ONNX artifact. This intermediate representation enables the model to be deployed across various runtimes (Python, C++, Julia) or, as implemented in this work, in JavaScript, directly into web browsers via the ONNX Web Runtime to power the Delphi App and Observable HQ.

## 3. Results

This paper details the development of a proof-of-concept web application using the Delphi 2M generative transformer model for an in-browser, privacy-preserving, health trajectory projection model. By leveraging the ONNX web runtime, the entire inference pipeline was shifted from the server to the client browser, creating a "build once and run anywhere" paradigm. We demonstrated that the same model powers both the user-facing web application, significantly improving its accessibility, and the ObservableHQ environment used for technical testing and validation. This enables the model to run efficiently on diverse platforms ranging from mobile devices to GPU-accelerated desktops. As a consequence, sensitive health trajectories input into and projected by the Delphi model remain exclusively on the user's local device. We also explored the role of the custom JavaScript SDK for transparent deployment, validating a high-performance engineering path for privacy-first generative AI applications.

**Software Resources**
Web application: https://epiverse.github.io/delphiTrajectories/
Github repository: https://github.com/epiverse/delphiTrajectories
SDK repository: https://github.com/episphere/delphi-onnx
SDK notebook: https://observablehq.com/@prafulb/delphi-2m-js-sdk

### 4. Discussion

This engineering exercise demonstrates a secure deployment of Generative AI models in privacy-sensitive domains. By successfully migrating the Delphi 2M inference pipeline to the client-side using ONNX, the model's predictive utility was fully decoupled from centralized data processing. The application's computational boundary resides entirely within the user's device, no individual health data is transmitted over the network for prediction, aligning the solution with strict data governance standards and placing user control at the core of the deployment. This work provides a concrete and positive case study for converting existing workflows to adhere more strongly to the Reusability (R) and Interoperability (I) components of the FAIR data principles.

While the technical architecture proves the viability of client-side inference, the predictive accuracy of the current proof-of-concept is limited. The model used for this demonstration was trained exclusively on a small subset of entirely synthetic data released by the authors of the original model, and therefore does not match the performance of the full model presented in the original work, which used 400k trajectories. Future models would use the full real-world dataset[22,23] to better demonstrate the potential clinical utility of the approach.

Finally, to address the challenge of future model updates without compromising the client-side privacy-preserving guarantee, we will explore integrating client-side fine-tuning, potentially by incorporating human feedback through active learning or more broadly via decentralized federated learning, allowing the model to learn and improve while sensitive data remains exclusively on the user's device.

### 5. Conclusion

The successful development of a web application utilizing the Delphi 2M generative transformer entirely client-side[24–28], as demonstrated by this prototype[18], validates a critical approach to deploying powerful AI in privacy-sensitive domains. By standardizing the model via ONNX and abstracting complexity through a dedicated SDK, this work establishes a secure, high-performance architecture for privacy-preserving computation[29–36].

**Acknowledgements**
The authors of this study are indebted to the team that developed Delphi 2M for the thorough and generous documentation: we were able to conceive and deploy a fully functional in-browser application without any other assistance than the documentation and libraries in their original report.